\documentclass[11pt,a4paper]{article}
\usepackage[hyperref]{naaclhlt2018}
\usepackage{times}
\usepackage{latexsym}
\usepackage{booktabs}
\usepackage{amsmath}
\usepackage{graphicx}
\usepackage{url}
\usepackage{caption}
\usepackage{subcaption}
\usepackage{bbm}

\definecolor{roamdarkblue}{HTML}{0499CC}
\definecolor{roamlightblue}{HTML}{03A9F4}
\definecolor{roamdarkgray}{HTML}{838A8A}
\definecolor{roamlightgray}{HTML}{B8B8B8}
\definecolor{roamgreen}{HTML}{4D8951}
\definecolor{roamblack}{HTML}{212121}
\definecolor{roamsteelblue}{HTML}{9BB8D7}
\definecolor{roamorange}{HTML}{FDBA58}
\definecolor{roamwhite}{HTML}{FAFAFA}
\definecolor{roampurple}{HTML}{876DB5}

\definecolor{superlightgray}{HTML}{DDDDDD}
\definecolor{superlightgreen}{HTML}{B4FFB4}
\definecolor{superlightorange}{HTML}{FFD090}

\newcommand{\roamdarkblue}[1]{{\color{roamdarkblue}#1}}

\aclfinalcopy 

\newcommand{\secref}[1]{sec.~\ref{#1}}

\newcommand{\Figref}[1]{Fig.~\ref{#1}}
\newcommand{\figref}[1]{fig.~\ref{#1}}

\newcommand{\Tabref}[1]{Tab.~\ref{#1}}
\newcommand{\tabref}[1]{tab.~\ref{#1}}

\newcommand{\mittensweight}{\mu}

\newcommand{\dtag}[1]{\roamdarkblue{/#1}}

\title{Mittens: An Extension of GloVe for Learning Domain-Specialized
  Representations}

\author{
  Nicholas Dingwall \\
  Roam Analytics \\
  \texttt{nick@roamanalytics.com} \\\And
  Christopher Potts \\
  Stanford University and Roam Analytics \\
  \texttt{cgpotts@stanford.edu}
}

\date{}

\begin{document}
\maketitle
\begin{abstract}
  We present a simple extension of the GloVe representation learning
  model that begins with general-purpose representations
  and updates them based on data from a specialized domain. We show
  that the resulting representations can lead to faster learning and
  better results on a variety of tasks.
\end{abstract}


\section{Introduction}\label{sec:intro}

Many NLP tasks have benefitted from the public availability of
general-purpose vector representations of words trained on enormous
datasets, such as those released by the GloVe
\citep{pennington-socher-manning:2014:EMNLP2014} and fastText
\citep{bojanowski2016enriching} teams. These representations, when
used as model inputs, have been shown to lead to faster learning and
better results in a wide variety of settings
\citep{erhan2009difficulty,erhan2010does,Cases:Luong:Potts:2017}.

However, many domains require more specialized representations but
lack sufficient data to train them from scratch. We address this
problem with a simple extension of the GloVe model
\citep{pennington-socher-manning:2014:EMNLP2014} that synthesizes
general-purpose representations with specialized data sets.  The
guiding idea comes from the retrofitting work of
\citet{Faruqui-etal:2015}, which updates a space of existing
representations with new information from a knowledge graph while also
staying faithful to the original space (see also
\citealt{yu2014improving,mrkvsic2016counter,pilehvar2016improved}). We
show that the GloVe objective is amenable to a similar retrofitting
extension. We call the resulting model `Mittens', evoking the idea
that it is `GloVe with a warm start' or a `warmer GloVe'.

Our hypothesis is that Mittens representations synthesize the
specialized data and the general-purpose pretrained representations in
a way that gives us the best of both. To test this, we conducted a
diverse set of experiments. In the first, we learn GloVe and Mittens
representations on IMDB movie reviews and test them on separate IMDB
reviews using simple classifiers. In the second, we learn our
representations from clinical text and apply them to a sequence
labeling task using recurrent neural networks, and to edge detection
using simple classifiers. These experiments support our hypothesis
about Mittens representations and help identify where they are most
useful.


\section{Mittens}

This section defines the Mittens objective. We first vectorize GloVe
to help reveal why it can be extended into a retrofitting model.


\subsection{Vectorizing GloVe}

\begin{table*}
  \centering
  \begin{tabular}[c]{l@{\hspace{38pt}}r r r@{\hspace{38pt}} r r r}
    \toprule
                   & \multicolumn{6}{c}{Vocabulary size} \\[0.5ex]
                   & \multicolumn{3}{c}{CPU} & \multicolumn{3}{c}{GPU} \\[0.5ex]
    Implementation            & 5K & 10K & 20K & 5K & 10K & 20K  \\
    \midrule
    Non-vectorized TensorFlow & $14.02$  &  $63.80$ & $252.65$  & $13.56$  & $55.51$ & $226.41$ \\
    Vectorized Numpy          & $1.48$   &  $7.35$  & $50.03$   & $-$      & $-$     & $-$ \\
    Vectorized TensorFlow     & $1.19$   &  $5.00$  & $28.69$   & $0.27$   & $0.95$  & $3.68$ \\
    Official GloVe            & $0.66$   &  $1.24$  & $3.50$    & $-$      & $-$     & $-$\\
    \bottomrule
  \end{tabular}
  \caption{Speed comparisons. The values are seconds per iteration,
    averaged over 10 iterations each on 5 simulated corpora that
    produced count matrices with about $10\%$ non-zero cells. Only the
    training step for each model is timed. The CPU experiments were
    done on a machine with a 3.1~GHz Intel Core i7 chip and 16~GB of
    memory, and the GPU experiments were done on machine with a 16~GB
    NVIDIA Tesla V100~GPU and 61~GB of memory. Dashes mark tests that
    aren't applicable because the implementation doesn't perform GPU
    computations.
  }
  \label{tab:speed}
\end{table*}

For a word $i$ from vocabulary $V$ occurring in the context of word
$j$, GloVe learns representations $w_{i}$ and $\widetilde{w}_{j}$
whose inner product approximates the logarithm of the probability of
the words' co-occurrence.  Bias terms $b_{i}$ and $\tilde{b}_{j}$
absorb the overall occurrences of $i$ and $j$. A weighting function
$f$ is applied to emphasize word pairs that occur frequently and
reduce the impact of noisy, low frequency pairs.  This results in the
objective
\begin{equation*}
  J = \sum_{i, j=1}^{V} f\left(X_{ij}\right)
  \left(w_i^\top \widetilde{w}_j + b_i + \tilde{b}_j - \log X_{ij}\right)^2
  \label{eq:glove}
\end{equation*}
where $X_{ij}$ is the co-occurrence of $i$ and $j$.  Since
$\log X_{ij}$ is only defined for $X_{ij} > 0$, the sum excludes
zero-count word pairs.  As a result, existing implementations of GloVe
use an inner loop to compute this cost and associated derivatives.

However, since $f(0) = 0$, the second bracket is irrelevant
whenever $X_{ij} = 0$, and so replacing $\log X_{ij}$ with
\begin{equation*}
    g(X_{ij}) = \begin{cases}
        k, & \text{for } X_{ij} = 0 \\
        \log(X_{ij}), & \text{otherwise}
        \end{cases}
\end{equation*}
(for any $k$) does not affect the objective and reveals that the
cost function can be readily vectorized as
\begin{equation*}
  J = f(X) M^\top M
  \label{eq:vec}
\end{equation*}
where $M=W^\top \widetilde{W} + b1^\top + 1 \tilde{b}^\top - g(X)$.
$W$ and $\widetilde{W}$ are matrices whose columns comprise the
word and context embedding vectors, and $g$ is applied elementwise.  Because
$f(X_{ij})$ is a factor of all terms of the derivatives, the gradients
are identical to the original GloVe implementation too.

To assess the practical value of vectorizing GloVe, we implemented the
model\footnote{{\scriptsize\url{https://github.com/roamanalytics/mittens}}}
in pure Python/Numpy \citep{Numpy:2011} and in TensorFlow
\citep{tensorflow2015-whitepaper}, and we compared these
implementations to a non-vectorized TensorFlow implementation and to
the official GloVe C implementation
\citep{pennington-socher-manning:2014:EMNLP2014}.\footnote{We also
  considered a non-vectorized Numpy implementation, but it was too
  slow to be included in our tests (a single iteration with a 5K
  vocabulary took 2 hrs 38 mins).}  The results of these tests are in
\tabref{tab:speed}. Though the C implementation is the fastest (and
scales to massive vocabularies), our vectorized TensorFlow
implementation is a strong second-place finisher, especially where GPU
computations are possible.


\subsection{The Mittens Objective Function}

This vectorized implementation makes it apparent that we can extend
GloVe into a retrofitting model by adding a term to the objective that
penalizes the squared euclidean distance from the learned embedding
$\widehat{w}_{i} = w_{i} + \widetilde{w}_{i}$ to an existing one,
$r_{i}$:
\begin{equation*}
  J_{\text{Mittens}} =
  J + \mittensweight \sum_{i \in R}
         \|\widehat{w}_{i} - r_{i}\|^{2}.
  \label{eq:mittens}
\end{equation*}
Here, $R$ contains the subset of words in the new vocabulary for which
prior embeddings are available (i.e., $R = V \cap V^\prime$ where
$V^\prime$ is the vocabulary used to generate the prior embeddings),
and $\mittensweight$ is a non-negative real-valued weight.  When
$\mittensweight = 0$ or $R$ is empty (i.e., there is no original
embedding), the objective reduces to GloVe's.

As in retrofitting, this objective encodes two opposing pressures:
the GloVe objective (left term), which favors changing representations,
and the distance measure (right term), which favors remaining true to
the original inputs. We can control this trade off by decreasing or
increasing $\mittensweight$.

In our experiments, we always begin with 50-dimensional `Wikipedia
2014 + Gigaword 5' GloVe
representations\footnote{{\scriptsize
\url{http://nlp.stanford.edu/data/glove.6B.zip}}} --
henceforth `External GloVe' -- but the model is compatible with any
kind of ``warm start''.


\subsection{Notes on Mittens Representations}\label{sec:notes}

\begin{figure}[t]
  \centering
  \begin{subfigure}{1\linewidth}
    \centering
    \includegraphics[width=1\textwidth]{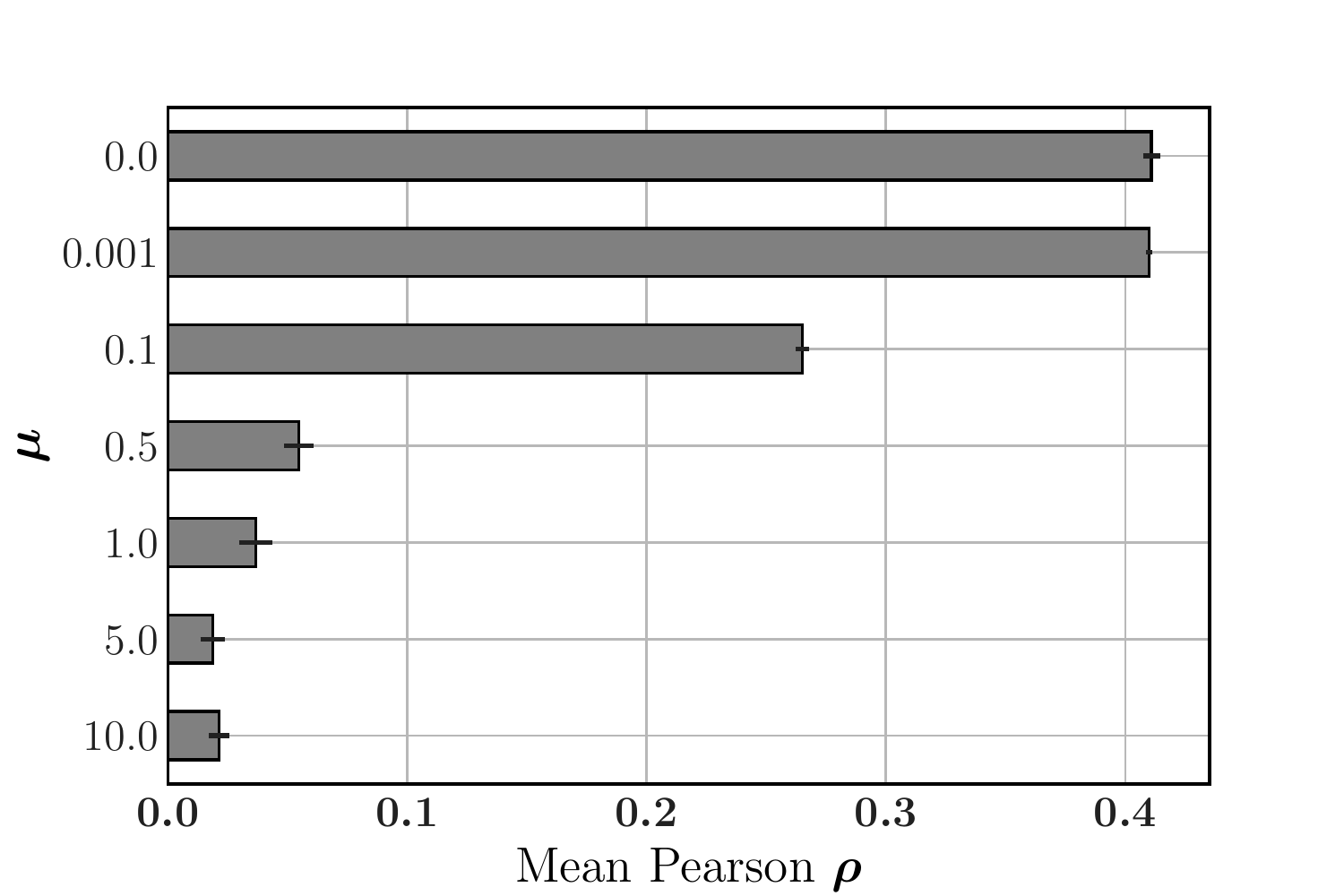}
    \caption{Correlations between the dot product of pairs of learned
      vectors and their log probabilities.}
    \label{fig:corr}
  \end{subfigure}

  \vspace{10pt}

  \begin{subfigure}{1\linewidth}
    \centering
    \includegraphics[width=1\textwidth]{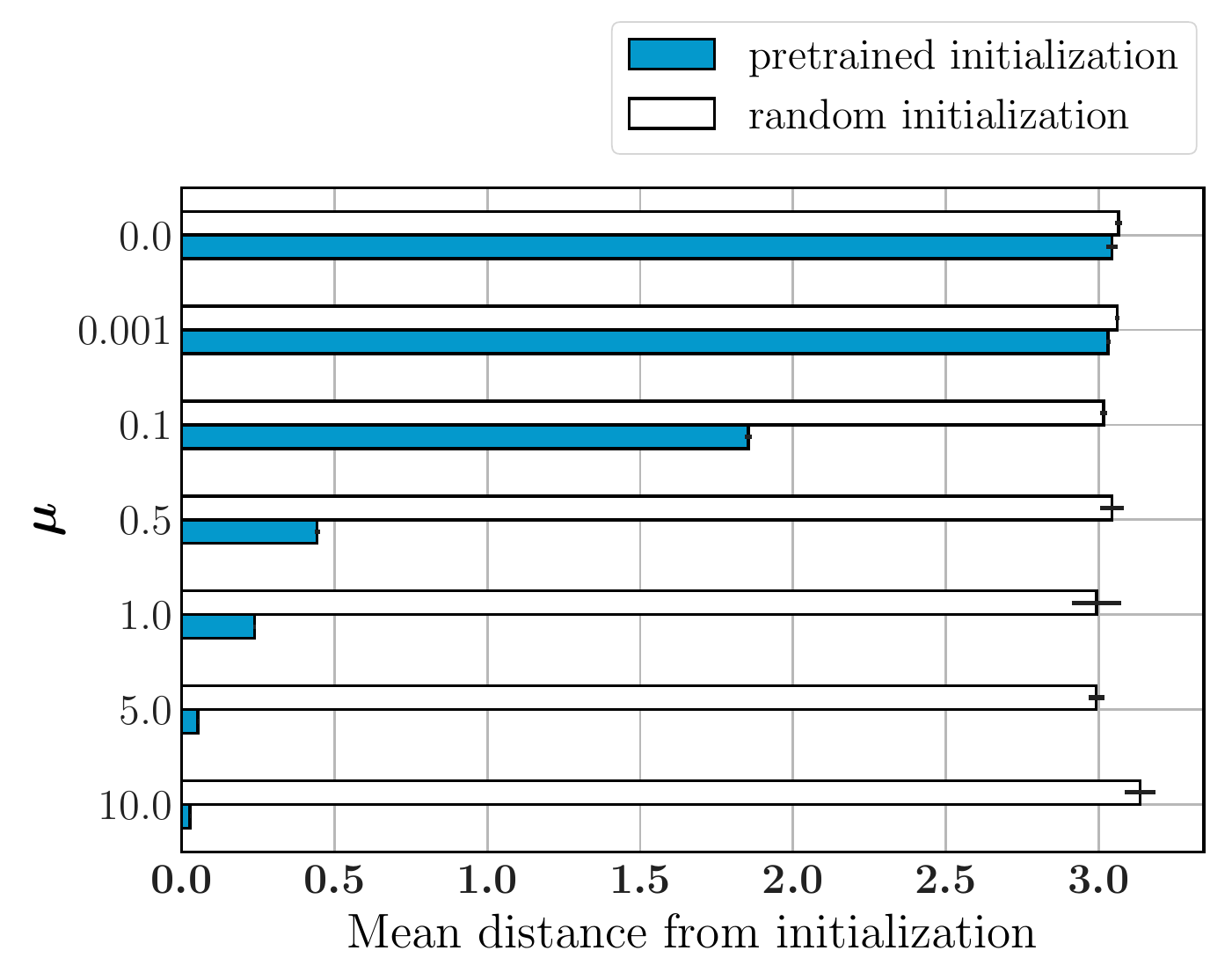}
    \caption{Distances between initial and learned embeddings, for
      words with and without pretrained initializations.  As $\mu$
      gets larger, the pressure to stay close to the original
      increases.}
    \label{fig:dist}
  \end{subfigure}
  \caption{Simulations assessing Mittens' faithfulness to the original
    GloVe objective and to its input embeddings. $\mittensweight=0$ is
    regular GloVe.}
  \label{fig:sim}
\end{figure}

GloVe's objective is that the log probability of words $i$ and $j$
co-occurring be proportional to the dot product of their learned
vectors. One might worry that Mittens distorts this, thereby
diminishing the effectiveness of GloVe. To assess this, we simulated
500-dimensional square count matrices and original embeddings for 50\%
of the words. Then we ran Mittens with a range of values of
$\mittensweight$. The results for five trials are summarized in
\figref{fig:sim}: for reasonable values of $\mittensweight$, the
desired correlation remains high (\figref{fig:corr}), even as vectors
with initial embeddings stay close to those inputs, as desired
(\figref{fig:dist}).

\section{Sentiment Experiments}\label{sec:sentiment}

For our sentiment experiments, we train our representations on the
unlabeled part of the IMDB review dataset released by
\citet{Maas-etal2011}. This simulates a common use-case: Mittens
should enable us to achieve specialized representations for these
reviews while benefiting from the large datasets used to train
External GloVe.

\begin{table}[t]
  \centering
  \begin{tabular}[t]{l r c}
    \toprule
    Representations & Accuracy & 95\% CI \\
    \midrule
    Random         & $62.00$ & $[61.28,62.53]$ \\
    External GloVe & $72.19$ & $-$ \\
    IMDB GloVE     & $76.38$ & $[75.76,76.72]$ \\
    Mittens        & $77.39$ & $[77.23,77.50]$ \\
    \bottomrule
  \end{tabular}
  \caption{IMDB test-set classification results. A difference of
    $1\%$ corresponds to $250$ examples. For all but `External GloVE',
    we report means (with bootstrapped confidence intervals)
    over five runs of creating the embeddings and cross-validating the
    classifier's hyperparameters, mainly to help verify that the
    differences do not derive from variation in the
    representation learning phase.
  }
  \label{tab:imdb-results}
\end{table}


\subsection{Word Representations}\label{sec:imdb-reps}

All our representations begin from a common count matrix obtained by
tokenizing the unlabeled movie reviews in a way that splits out
punctuation, downcases words unless they are written in all uppercase,
and preserves emoticons and other common social media mark-up. We say
word $i$ co-occurs with word $j$ if $i$ is within 10 words
to the left or right of $j$, with the counts weighted by $1/d$
where $d$ is the distance in words from $j$. Only words with at
least 300 tokens are included in the matrix, yielding a vocabulary of
3,133 words.

For regular GloVe representations derived from the IMDB data --
`IMDB GloVE' -- we train 50-dimensional representations and use the
default parameters from
\citealt{pennington-socher-manning:2014:EMNLP2014}: \mbox{$\alpha = 0.75$},
$x_{\max} = 100$, and a learning rate of $0.05$. We optimize with
AdaGrad \citep{Duchi:Hazan:Singer:2011}, also as in the original
paper, training for 50K epochs.

For Mittens, we begin with External GloVe.  The few words in the IMDB
vocabulary that are not in this GloVe vocabulary receive random
initializations with a standard deviation that matches that of the
GloVe representations. Informed by our simulations, we train
representations with the Mittens weight $\mittensweight = 0.1$. The
GloVe hyperparameters and optimization settings are as
above. Extending the correlation analysis of \figref{fig:corr} to
these real examples, we find that the GloVe representations generally
have Pearson's $\rho \approx 0.37$, Mittens $\rho \approx 0.47$.
We speculate that the improved correlation is due to the low-variance
external GloVe embedding smoothing out noise from our co-occurrence
matrix.


\newcommand{\other}{\dtag{O} }
\newcommand{\positive}{\dtag{D} }
\newcommand{\concern}{\dtag{C} }
\newcommand{\ruledout}{\dtag{R} }

\begin{table*}[t]
  \centering
  \begin{subtable}{1\linewidth}
    \centering
    \begin{tabular}[t]{r@{. }l}
      \toprule
      1 & No\other eye\ruledout pain\ruledout or\other eye\ruledout discharge\ruledout .\other \\
      2 & Asymptomatic\positive bacteriuria\positive ,\other could\other be\other neurogenic\concern bladder\concern disorder\concern .\other \\
      3 & Small\concern embolism\concern in\concern either\concern lung\concern cannot\other be\other excluded\other .\other \\
      \bottomrule
    \end{tabular}
    \caption{Short disease diagnosis labeled examples.
      `O': `Other';
      `D': `Positive Diagnosis';
      `C': `Concern';
      `R': `Ruled Out'.}
    \label{tab:diagnosis-examples}
  \end{subtable}
  \caption{Disease diagnosis examples.}
  \label{tab:diagnosis}
\end{table*}

\begin{figure*}[t]
  \centering
  \includegraphics[width=1\linewidth]{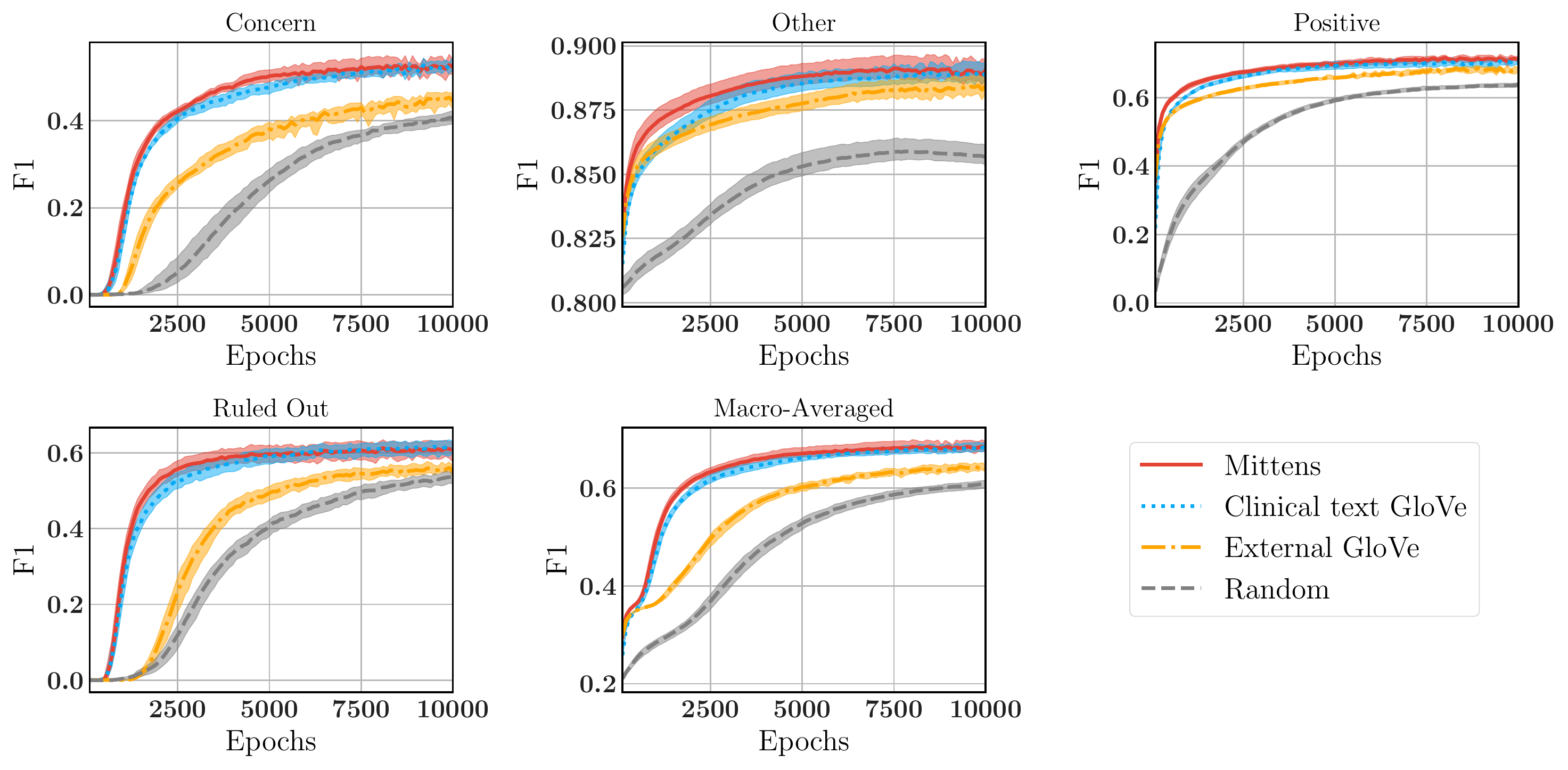}
  \caption{Disease diagnosis test-set accuracy as a function of
    training epoch, with bootstrapped confidence intervals.
    Mitten learns fastest for all categories.}
  \label{fig:diagnosis-curve}
\end{figure*}

\subsection{IMDB Sentiment Classification}\label{sec:sentiment-classify}

The labeled part of the IMDB sentiment dataset defines a
positive/negative classification problem with 25K labeled reviews for
training and 25K for testing. We represent each review by the
element-wise sum of the representation of each word in the review, and
train a random forest model \citep{Ho:1995,Breiman:2001} on these
representations. The rationale behind this experimental set-up is that
it fairly directly evaluates the vectors themselves; whereas the
neural networks we evaluate next can update the representations, this
model relies heavily on their initial values.

Via cross-validation on the training data, we optimize the number of
trees, the number of features at each split, and the maximum depth of
each tree. To help factor out variation in the representation learning
step \citep{reimers2017reporting}, we report the average accuracies
over five separate complete experimental runs.

Our results are given in \tabref{tab:imdb-results}. Mittens
outperforms External GloVe and IMDB GloVe, indicating that it
effectively combines complementary information from both.


\section{Clinical Text Experiments}

Our clinical text experiments begin with 100K clinical notes
(transcriptions of the reports healthcare providers create summarizing
their interactions with patients during appointments) from Real Health
Data.\footnote{{\scriptsize\url{http://www.realhealthdata.com}}} These
notes are divided into informal segments that loosely follow the
`SOAP' convention for such reporting (Subjective, Objective,
Assessment, Plan). The sample has 1.3 million such segments, and these
segments provide our notion of `document'.


\subsection{Word Representations}

The count matrix is created from the clinical text using the
specifications described in \secref{sec:imdb-reps}, but with the count
threshold set to 500 to speed up optimization. The final matrix has a
6,519-word vocabulary. We train Mittens and GloVe as in
\secref{sec:imdb-reps}. The correlations in the sense of
\figref{fig:corr} are $\rho \approx 0.51$ for both GloVe and Mittens.

\begin{table*}[t]
  \begin{subtable}[t]{0.28\textwidth}
    \centering
    \setlength{\tabcolsep}{2pt}
    \begin{tabular}[t]{l *{2}{r} }
      \toprule
      Subgraph  & Nodes & Edges \\
      \midrule
      disorder  & $72,551$ & $408,411$ \\
      procedure & $53,616$ & $264,000$ \\
      finding   & $35,544$ & $76,563$ \\
      organism  & $33,721$ & $41,090$ \\
      substance & $26,207$ & $46,333$ \\
      \bottomrule
    \end{tabular}
    \caption{Subgraph sizes.}
    \label{tab:snomed-subgraphs}
  \end{subtable}
  \hfill
  \begin{subtable}[t]{0.7\textwidth}
  \centering
  \setlength{\tabcolsep}{2pt}
  \begin{tabular}[t]{l *{5}{r} }
    \toprule
    Representations      & disorder         & procedure        & finding          & organism         & substance \\
    \midrule
    Random               & $56.05$          & $55.97$          &
                                                                 $75.14$          & $68.15$          & $64.72$ \\
    External GloVe       & $\mathbf{69.31}$ & $65.89$          & $\mathit{80.72}$ & $74.12$          & $\mathbf{77.58}$ \\
    Clinical text GloVe  & $66.19$          & $64.96$          & $79.18$          & $73.42$          & $73.37$ \\
    Mittens              & $67.59$          & $\mathbf{66.59}$ & $\mathbf{80.74}$ & $\mathbf{74.53}$ & $76.51$ \\
    \bottomrule
  \end{tabular}
  \caption{Mean macro-F1 by subgraph (averages from 10 random train/test
    splits). Italics mark systems for which $p \geq 0.05$ in a
    comparison with the top system numerically, according to a
    Wilcoxon signed-rank test.}
  \label{tab:snomed-results}
  \end{subtable}
  \caption{SNOMED subgraphs and results. For the `disorder' graph (the largest), a
    difference of $0.1\%$ corresponds to $408$ examples. For the
    `substance' graph (the smallest), it corresponds to $46$ examples.}
\end{table*}

\subsection{Disease Diagnosis Sequence Modeling}

Here we use a recurrent neural network (RNN) to evaluate our
representations. We sampled 3,206 sentences from clinical texts
(disjoint from the data used to learn word representations) containing
disease mentions, and labeled these mentions as `Positive diagnosis',
`Concern', `Ruled Out', or `Other'.  \Tabref{tab:diagnosis-examples}
provides some examples. We treat this as a sequence labeling problem,
using `Other' for all unlabeled tokens. Our RNN has a single
50-dimensional hidden layer with LSTM cells
\citep{Hochreiter:Schmidhuber:1997}, and the inputs are updated during
training.

\Figref{fig:diagnosis-curve} summarizes the results of these
experiments based on 10 random train/test with 30\% of the sentences
allocated for testing. Since the inputs can be updated, we expect all the
initialization schemes to converge to approximately the same performance
eventually (though this seems not to be the case in practical terms
for Random or External GloVE). However, Mittens learns fastest for all
categories, reinforcing the notion that Mittens is a sensible default
choice to leverage both domain-specific and large-scale data.


\subsection{SNOMED CT edge prediction}

Finally, we wished to see if Mittens representations would generalize
beyond the specific dataset they were trained on. SNOMED CT is a
public, widely-used graph of healthcare concepts and their
relationships \citep{spackman1997snomed}. It contains 327K nodes,
classified into 169 semantic types, and 3.8M edges. Our clinical notes
are more colloquial than SNOMED's node names and cover only some of
its semantic spaces, but the Mittens representations should still be
useful here.

For our experiments, we chose the five largest semantic types;
\tabref{tab:snomed-subgraphs} lists these subgraphs along with their
sizes. Our task is edge prediction: given a pair of nodes in a
subgraph, the models predict whether there should be an edge between
them. We sample 50\% of the non-existent edges to create a balanced
problem. Each node is represented by the sum of the vectors for the
words in its primary name, and the classifier is trained on the
concatenation of these two node representations. To help assess
whether the input representations truly generalize to new cases, we
ensure that the sets of nodes seen in training and testing are
disjoint (which entails that the edge sets are disjoint as well), and
we train on just 50\% of the nodes. We report the results of ten
random train/test splits.

The large scale of these problems prohibits the
large hyperparameter search described in
\secref{sec:sentiment-classify}, so we used the best settings
from those experiments (500 trees per forest, square root of the total
features at each split, no depth restrictions).

Our results are summarized in \tabref{tab:snomed-results}. Though the
differences are small numerically, they are meaningful because of the
large size of the graphs (\tabref{tab:snomed-subgraphs}). Overall,
these results suggest that Mittens is at its best where there is a
highly-specialized dataset for learning representations, but that it
is a safe choice even when seeking to transfer the representations to
a new domain.


\section{Conclusion}

We introduced a simple retrofitting-like extension to the original
GloVe model and showed that the resulting representations were
effective in a number of tasks and models, provided a substantial
(unsupervised) dataset in the same domain is available to tune the
representations.  The most natural next step would be to study similar
extensions of other representation-learning models.


\section{Acknowledgements}

We thank Real Health Data for providing our clinical texts,
Ben Bernstein,
Andrew Maas,
Devini Senaratna,
and
Kevin Reschke
for valuable comments and discussion,
and Grady Simon for making his Tensorflow implementation
of GloVe available \citep{GradySimon:2017}.

\bibliographystyle{acl_natbib}
\bibliography{mittens-bib}

\begin{thebibliography}{19}
\expandafter\ifx\csname natexlab\endcsname\relax\def\natexlab#1{#1}\fi

\bibitem[{Abadi et~al.(2015)Abadi, Agarwal, Barham, Brevdo, Chen, Citro,
  S.~Corrado, Davis, Dean, Devin, Ghemawat, Goodfellow, Harp, Irving, Isard,
  Jozefowicz, Rafal, Kaiser, Kudlur, Levenberg, Man\'{e}, Monga, Moore, Murray,
  Olah, Schuster, Shlens, Steiner, Sutskever, Talwar, Tucker, Vanhoucke,
  Vasudevan, Vi\'{e}gas, Vinyals, Warden, Wattenberg, Wicke, Yu, and
  Xiaoqiang}]{tensorflow2015-whitepaper}
Mart\'{\i}n Abadi, Ashish Agarwal, Paul Barham, Eugene Brevdo, Zhifeng Chen,
  Craig Citro, Greg S.~Corrado, Andy Davis, Jeffrey Dean, Matthieu Devin,
  Sanjay Ghemawat, Ian Goodfellow, Andrew Harp, Geoffrey Irving, Michael Isard,
  Yangqing~Jia Jozefowicz, Rafal, Lukasz Kaiser, Manjunath Kudlur, Josh
  Levenberg, Dandelion Man\'{e}, Rajat Monga, Sherry Moore, Derek Murray, Chris
  Olah, Mike Schuster, Jonathon Shlens, Benoit Steiner, Ilya Sutskever, Kunal
  Talwar, Paul Tucker, Vincent Vanhoucke, Vijay Vasudevan, Fernanda Vi\'{e}gas,
  Oriol Vinyals, Pete Warden, Martin Wattenberg, Martin Wicke, Yuan Yu, and
  Zheng Xiaoqiang. 2015.
\newblock \href {https://www.tensorflow.org} {{TensorFlow}: Large-scale machine
  learning on heterogeneous systems}.

\bibitem[{Bojanowski et~al.(2016)Bojanowski, Grave, Joulin, and
  Mikolov}]{bojanowski2016enriching}
Piotr Bojanowski, Edouard Grave, Armand Joulin, and Tomas Mikolov. 2016.
\newblock \href {https://arxiv.org/abs/1607.04606} {Enriching word vectors with
  subword information}.
\newblock ArXiv:1607.04606.

\bibitem[{Breiman(2001)}]{Breiman:2001}
Leo Breiman. 2001.
\newblock Random forests.
\newblock \emph{Machine learning}, 45(1):5--32.

\bibitem[{Cases et~al.(2017)Cases, Luong, and Potts}]{Cases:Luong:Potts:2017}
Ignacio Cases, Minh-Thang Luong, and Christopher Potts. 2017.
\newblock \href {https://arxiv.org/abs/1710.02076} {On the effective use of
  pretraining for natural language inference}.
\newblock ArXiv:1710.02076.

\bibitem[{Duchi et~al.(2011)Duchi, Hazan, and Singer}]{Duchi:Hazan:Singer:2011}
John Duchi, Elad Hazan, and Yoram Singer. 2011.
\newblock \href {http://jmlr.csail.mit.edu/papers/v12/duchi11a.html} {Adaptive
  subgradient methods for online learning and stochastic optimization}.
\newblock \emph{Journal of Machine Learning Research}, pages 2121--2159.

\bibitem[{Erhan et~al.(2010)Erhan, Bengio, Courville, Manzagol, Vincent, and
  Bengio}]{erhan2010does}
Dumitru Erhan, Yoshua Bengio, Aaron Courville, Pierre-Antoine Manzagol, Pascal
  Vincent, and Samy Bengio. 2010.
\newblock \href {http://jmlr.org/papers/v11/erhan10a.html} {Why does
  unsupervised pre-training help deep learning?}
\newblock \emph{The Journal of Machine Learning Research}, 11:625--660.

\bibitem[{Erhan et~al.(2009)Erhan, Manzagol, Bengio, Bengio, and
  Vincent}]{erhan2009difficulty}
Dumitru Erhan, Pierre-Antoine Manzagol, Yoshua Bengio, Samy Bengio, and Pascal
  Vincent. 2009.
\newblock \href {http://proceedings.mlr.press/v5/erhan09a.html} {The difficulty
  of training deep architectures and the effect of unsupervised pre-training}.
\newblock In \emph{International Conference on Artificial Intelligence and
  Statistics}, pages 153--160.

\bibitem[{Faruqui et~al.(2015)Faruqui, Dodge, Jauhar, Dyer, Hovy, and
  Smith}]{Faruqui-etal:2015}
Manaal Faruqui, Jesse Dodge, Sujay~Kumar Jauhar, Chris Dyer, Eduard Hovy, and
  Noah~A. Smith. 2015.
\newblock \href {http://www.aclweb.org/anthology/N15-1184} {Retrofitting word
  vectors to semantic lexicons}.
\newblock In \emph{Proceedings of the 2015 Conference of the North American
  Chapter of the Association for Computational Linguistics: Human Language
  Technologies}, pages 1606--1615, Stroudsburg, PA. Association for
  Computational Linguistics.

\bibitem[{Ho(1995)}]{Ho:1995}
Tin~Kam Ho. 1995.
\newblock Random decision forests.
\newblock In \emph{Proceedings of the Third International Conference on
  Document Analysis and Recognition}, volume~1, pages 278--282. IEEE.

\bibitem[{Hochreiter and Schmidhuber(1997)}]{Hochreiter:Schmidhuber:1997}
Sepp Hochreiter and J{\"u}rgen Schmidhuber. 1997.
\newblock Long short-term memory.
\newblock \emph{Neural Computation}, 9(8):1735--1780.

\bibitem[{Maas et~al.(2011)Maas, Daly, Pham, Huang, Ng, and
  Potts}]{Maas-etal2011}
Andrew~L. Maas, Raymond~E. Daly, Peter~T. Pham, Dan Huang, Andrew~Y. Ng, and
  Christopher Potts. 2011.
\newblock \href {http://www.aclweb.org/anthology/P11-1015} {Learning word
  vectors for sentiment analysis}.
\newblock In \emph{Proceedings of the 49th Annual Meeting of the {A}ssociation
  for {C}omputational {L}inguistics}, pages 142--150, Portland, Oregon.
  Association for Computational Linguistics.

\bibitem[{Mrk{\v{s}}i{\'{c}} et~al.(2016)Mrk{\v{s}}i{\'{c}},
  {\'O}~S{\'e}aghdha, Thomson, Ga{\v{s}}i{\'{c}}, Rojas-Barahona, Su, Vandyke,
  Wen, and Young}]{mrkvsic2016counter}
Nikola Mrk{\v{s}}i{\'{c}}, Diarmuid {\'O}~S{\'e}aghdha, Blaise Thomson, Milica
  Ga{\v{s}}i{\'{c}}, Lina~M. Rojas-Barahona, Pei-Hao Su, David Vandyke,
  Tsung-Hsien Wen, and Steve Young. 2016.
\newblock \href {https://doi.org/10.18653/v1/N16-1018} {Counter-fitting word
  vectors to linguistic constraints}.
\newblock In \emph{Proceedings of the 2016 Conference of the North American
  Chapter of the Association for Computational Linguistics: Human Language
  Technologies}, pages 142--148. Association for Computational Linguistics.

\bibitem[{Pennington et~al.(2014)Pennington, Socher, and
  Manning}]{pennington-socher-manning:2014:EMNLP2014}
Jeffrey Pennington, Richard Socher, and Christopher~D. Manning. 2014.
\newblock \href {http://www.aclweb.org/anthology/D14-1162} {{GloVe}: Global
  vectors for word representation}.
\newblock In \emph{Proceedings of the 2014 Conference on Empirical Methods in
  Natural Language Processing (EMNLP)}, pages 1532--1543, Doha, Qatar.
  Association for Computational Linguistics.

\bibitem[{Pilehvar and Collier(2016)}]{pilehvar2016improved}
Mohammad~Taher Pilehvar and Nigel Collier. 2016.
\newblock \href {https://doi.org/10.18653/v1/W16-2902} {Improved semantic
  representation for domain-specific entities}.
\newblock In \emph{Proceedings of the 15th Workshop on Biomedical Natural
  Language Processing}, pages 12--16. Association for Computational
  Linguistics.

\bibitem[{Reimers and Gurevych(2017)}]{reimers2017reporting}
Nils Reimers and Iryna Gurevych. 2017.
\newblock \href {https://arxiv.org/pdf/1707.09861.pdf} {Reporting score
  distributions makes a difference: Performance study of {LSTM}-networks for
  sequence tagging}.
\newblock \emph{arXiv:1707.09861}.

\bibitem[{Simon(2017)}]{GradySimon:2017}
Grady Simon. 2017.
\newblock \href {https://github.com/GradySimon/tensorflow-glove} {An
  implementation of {GloVe} in {TensorFlow}}.

\bibitem[{Spackman et~al.(1997)Spackman, Campbell, and
  C{\^o}t{\'e}}]{spackman1997snomed}
Kent~A Spackman, Keith~E Campbell, and Roger~A C{\^o}t{\'e}. 1997.
\newblock {SNOMED RT}: A reference terminology for health care.
\newblock In \emph{Proceedings of the AMIA Annual Fall Symposium}, page 640.
  American Medical Informatics Association.

\bibitem[{van~der Walt et~al.(2011)van~der Walt, Colbert, and
  Varoquaux}]{Numpy:2011}
St{\'e}fan van~der Walt, S.~Chris Colbert, and Ga{\"e}l Varoquaux. 2011.
\newblock The {NumPy} array: A structure for efficient numerical computation.
\newblock \emph{Computing in Science and Engineering}, 13:22--30.

\bibitem[{Yu and Dredze(2014)}]{yu2014improving}
Mo~Yu and Mark Dredze. 2014.
\newblock \href {https://doi.org/10.3115/v1/P14-2089} {Improving lexical
  embeddings with semantic knowledge}.
\newblock In \emph{Proceedings of the 52nd Annual Meeting of the Association
  for Computational Linguistics (Volume 2: Short Papers)}, pages 545--550.
  Association for Computational Linguistics.

\end{thebibliography}

\end{document}